\documentclass[pdflatex,sn-basic,Numbered]{sn-jnl}

\usepackage{microtype}
\usepackage{graphicx}%
\usepackage{multirow}%
\usepackage{amsmath,amssymb,amsfonts}%
\usepackage{amsthm}%
\usepackage{mathrsfs}%
\usepackage[title]{appendix}%
\usepackage{xcolor}%
\usepackage{textcomp}%
\usepackage{manyfoot}%
\usepackage{booktabs}%
\usepackage{algorithm}%
\usepackage{algorithmicx}%
\usepackage{algpseudocode}%
\usepackage{listings}%
\usepackage{csquotes} 
\usepackage[acronym]{glossaries}
\usepackage{cleveref}
\glsdisablehyper
\usepackage{svg}
\usepackage{svg-extract}

\theoremstyle{thmstyleone}%
\theoremstyle{thmstyletwo}%

\theoremstyle{thmstylethree}%

\begin{document}
\definecolor{pink}{HTML}{bd3786}
\definecolor{lila}{HTML}{481467}
\definecolor{blue}{HTML}{31688e}
\definecolor{green}{HTML}{bade28}

\definecolor{medium-risk}{HTML}{238a8d}
\definecolor{ensemble}{HTML}{56C667}


\newacronym{GDE}{GDE}{Generalization Disagreement Equality}
\newacronym{CP}{CP}{Clopper-Pearson}
\newacronym{APPA}{APPA}{Adjusted Pairwise Prediction Agreement}

\title[Article Title]{On Arbitrary Predictions from Equally Valid Models}

\author*[1,2]{\fnm{Sarah} \sur{Lockfisch}}\email{sarah.lockfisch@hpi.de}
\author[2]{\fnm{Kristian} \sur{Schwethelm}}
\author[2,3,4]{\fnm{Martin}  \sur{Menten}}
\author[5]{\fnm{Rickmer} \sur{Braren}}
\author[2,3,4]{\fnm{Daniel} \sur{Rückert}}
\author[2]{\fnm{Alexander} \sur{Ziller}}\equalcont{These authors contributed equally to this work.}
\author*[1]{\fnm{Georgios} \sur{Kaissis}}
\email{georg.kaissis@hpi.de}\equalcont{These authors contributed equally to this work.}

\affil[1]{Hasso Plattner Institute, University of Potsdam, Germany}
\affil[2]{Chair for AI in Healthcare and Medicine, Technical University of Munich, Germany}
\affil[3]{BioMedIA, Imperial College London, United Kingdom}
\affil[4]{Munich Center for Machine Learning, Germany}
\affil[5]{Universitätsklinikum Hamburg-Eppendorf, Germany}

\abstract{
Model multiplicity describes the existence of multiple models that fit the data equally well but can produce different predictions for individual samples, so-called predictive multiplicity.
In medicine, these models can admit conflicting predictions for the same patient---a risk that is poorly understood and insufficiently addressed.

In this study, we empirically analyze predictive multiplicity across multiple medical tasks and model architectures, and show practical strategies to mitigate it.
Our analysis reveals that (1) standard validation metrics fail to identify a uniquely optimal model.
(2) Models with statistically indistinguishable performance show variability in patient-level predictions, resulting in arbitrary and potentially harmful outcomes under any single model. 
However, predictive multiplicity does not affect samples equally, and the converse can be used to reduce predictive multiplicity.
We find that (3) high model capacity decreases predictive multiplicity by improving accuracy.
(4) Ensembles with an abstention strategy further enhance expected per-sample accuracy and stability.

Together, these findings highlight that predictive multiplicity is not merely a theoretical curiosity but a pervasive and practically significant issue in medical AI. 
We argue that accounting for multiplicity should be considered a core component of model evaluation and deployment in safety-critical domains.
}


\maketitle
\section{Introduction and Prior Work}\label{introduction} 
\begin{figure}[htb]
    \centering
    \includegraphics[width=\textwidth]{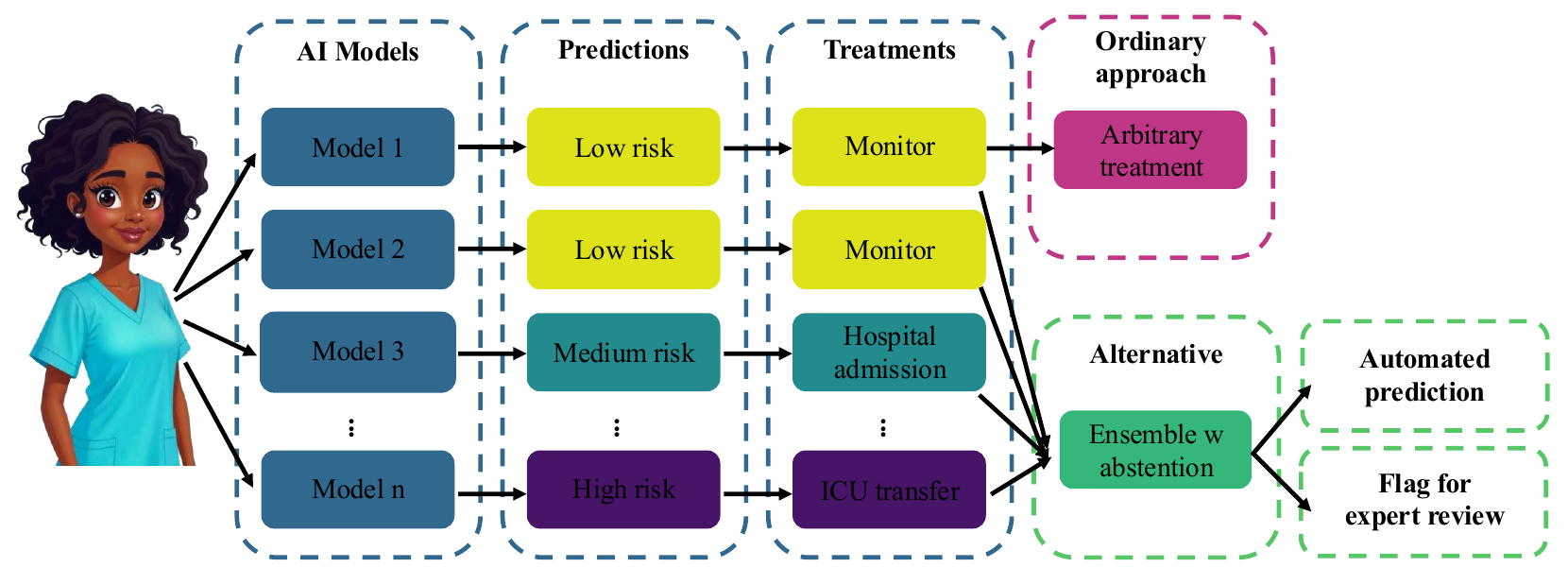}
    \caption{
    \textbf{
    Model multiplicity in clinical decision-making.}
    Multiple machine learning models achieve similar on-average performance and are thus equally valid choices to evaluate a patient's data (\textcolor{blue}{model 1 to n}). 
    However, these models may produce different risk predictions (\textcolor{green}{low}, \textcolor{medium-risk}{medium}, \textcolor{lila}{high}) for the same patient, leading to divergent treatment recommendations (\textcolor{green}{monitor}, \textcolor{medium-risk}{hospital admission}, \textcolor{lila}{ICU transfer}).
    As a result, a patient's care pathway can vary substantially---ultimately resulting in an \textcolor{pink}{arbitrary treatment}---depending on which model is used. 
    Rather than relying on a \textit{single} model, we can adopt an \textcolor{ensemble}{ensemble}-based strategy: cases with high inter-model agreement between models may be amenable to automated prediction, while those with low agreement should be reviewed by experts.}
    \label{fig:teaser}
\end{figure}

\textit{Model multiplicity} \cite{black2022model} describes the existence of many \textit{plausible} models for the same dataset without a principled way to determine a single optimal model, reflecting the underlying uncertainty in how the data should be interpreted.
\footnote{
In theory, the Bayes optimal predictor represents the optimal solution, i.e., the model that minimizes the expected loss with full knowledge of the data-generating distribution. 
However, this ideal predictor is not attainable in practice, as the true underlying distribution is unknown, and only finite data is available.}
In practice, multiple machine learning models can fit the same data equally well according to a performance metric (e.g., loss or accuracy), but may differ in their internal structure (e.g., the value of their parameters) and, more critically, in their individual predictions. 
Yet, for deployment, typically a \textit{single} model is chosen---commonly, without consideration for other, equally valid options. 
Using such an arbitrary\footnote{
In contrast, selecting the/ a single best model would be non-arbitrary from the perspective of instrumental
rationality: the model that performs as well as it can according to its own metric of success \cite{citron2014scored}.
However, model multiplicity highlights the existence of multiple equally optimal models that result in different predictions.
Thus, there is no self-interested reason to choose one optimal model over another \cite{creel2022algorithmic}.}
model is particularly problematic in high-stakes scenarios when other, equally well-performing models exist that produce different predictions on the same data point(s). 
If such a model is deployed in a clinical setting, \textit{a patient’s diagnosis---and their treatment---may ultimately depend on the choice of this specific model rather than on relevant properties of the patient’s data}.
This raises critical concerns about the justification for deploying this model in practice.
We illustrate the problem in \autoref{fig:teaser}.

The phenomenon of model multiplicity is not novel and has appeared under various names in literature, \textit{inter alia}, as the Rashomon Effect \cite{breiman2001statistical}, underspecification \cite{d2022underspecification}, or instability \cite{riley2023stability}.
Prior work has discussed both its opportunities \cite{rudin2024position} and challenges, notably \textit{predictive multiplicity} \cite{marx2020predictive}, where equally valid models produce conflicting predictions. 
While predictive multiplicity may be negligible in low-stakes settings or beneficial in certain contexts (e.g., to avoid systemic exclusion \cite{creel2022algorithmic}), it poses a serious risk in medicine, where model predictions inform counseling, resource allocation, and clinical care.
Conflicting predictions from equally valid models can erode trust, cause inconsistent treatment, and ultimately harm patients. 
Despite its relevance, systematic studies of predictive multiplicity in medicine remain scarce. 
Existing work identifies underspecification as a general source of instability \cite{d2022underspecification} without addressing its impact on individual predictions, or propose bootstrapping approaches \cite{riley2023stability, riley2023clinical} which are largely infeasible for modern ML; see \autoref{app:related-work} for details. 

In this study, we address this gap through a comprehensive empirical analysis of predictive multiplicity across multiple medical modalities (abdominal CTs, blood cell images, breast ultrasounds, OCT scans, and X-rays) and deep learning architectures (ConvNeXt, EfficientNet, GC ViT, ResNet50).
We find that recognizing and addressing model multiplicity not only exposes limitations of the \enquote{single model} paradigm but also offers a means to improve predictive stability and accuracy by levering inter-model (dis)agreement. 
Our large-scale study, based on the training of 1,500 models, leads to the following conclusions that form the core contributions of our work:
\begin{enumerate}
    \item \textbf{Validation performance is an unreliable indicator of generalization.}
    Thus, selecting any single, seemingly best-performing model becomes an arbitrary choice.
    \item \textbf{Relying on a single model exposes some patients to arbitrary predictions.}
    Yet, the underlying systematics can be exploited to improve stability and accuracy. 
    \item \textbf{Higher model capacity that improves accuracy reduces predictive multiplicity.}
    Accuracy maximization helps to minimize arbitrary predictions.
    \item \textbf{Ensembles with abstention reduce predictive multiplicity and improve accuracy.}
    Predictions with high inter-model consensus may be amenable for automation; insufficient consensus prompts deferral to human experts.
\end{enumerate}

\section{Results}\label{results}
Deployment typically relies on a single, \enquote{optimal} model, without considering other, equally valid alternatives. 
We instead examine the \textit{Rashomon set} \cite{breiman2001statistical}: the set of models that fit the data equally well and thus represent equally plausible solutions. 
Following prior work \cite{d2022underspecification, black2022selective}, we explore this set empirically by randomizing the initialization of model weights. 
Specifically, we replace the classification head of an ImageNet-pretrained model with a randomly initialized one and train the entire model. 
For each dataset/architecture combination we trained 50 model instances, which differ only in the weight initialization of the last layer while keeping all other components fixed.
This results in a total 1,500 models---1,000 for our main experiments and 500 to analyze model capacity.
In brief, our experiments cover aforementioned medical imaging datasets and model architectures and perform competitively (and often surpass) prior benchmarks (see \Cref{sec:methods} for details on datasets, architectures, training procedure, and performance). 

\subsection{The illusion of an optimal model}\label{subsec:rashomon-set}
\begin{figure}[!h]
    \centering
    \includegraphics[width=0.95\textwidth]{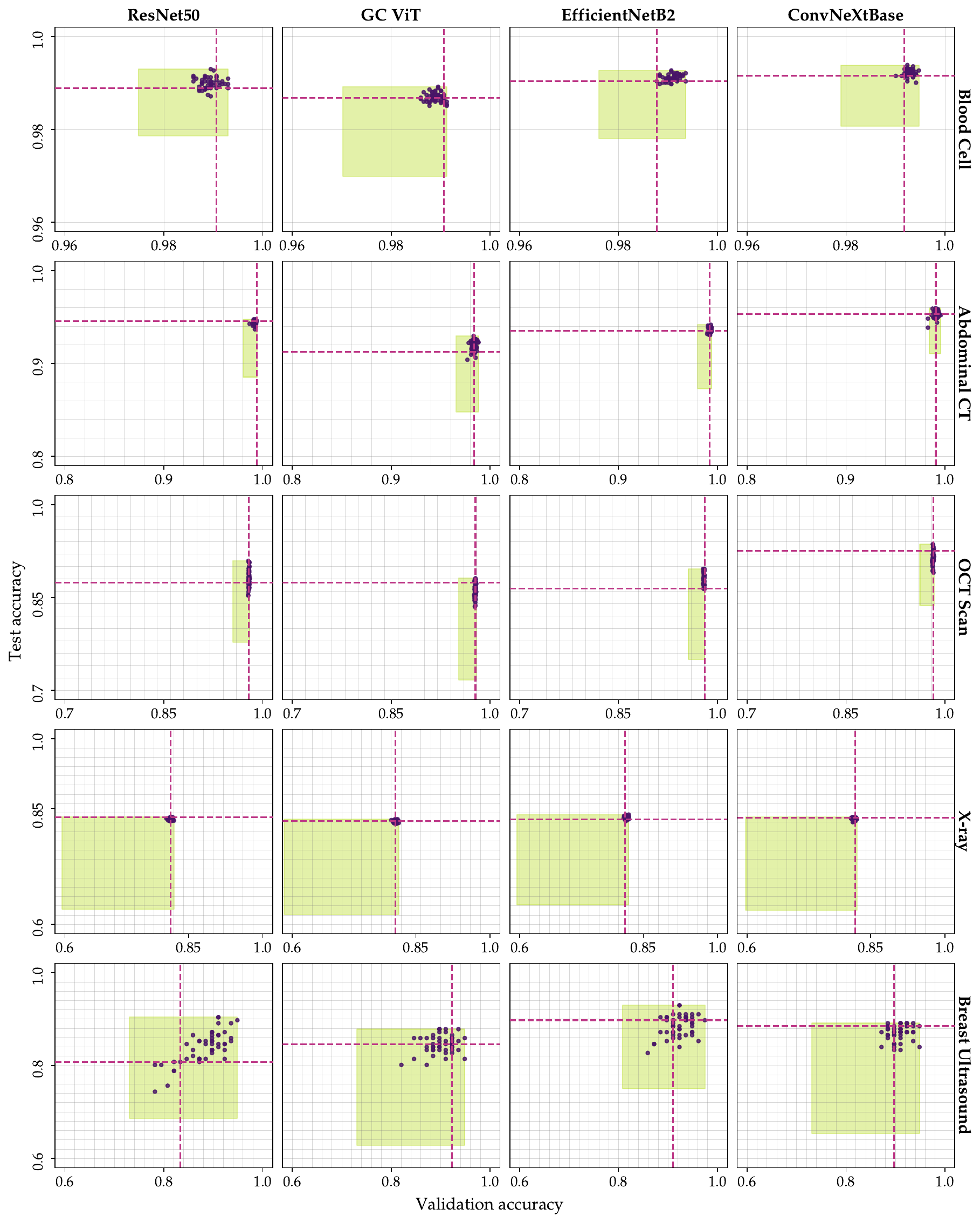}
    \caption{
    \textbf{Variation in accuracy within the empirical Rashomon set.}
    Each plot corresponds to a dataset/architecture combination with points (\textcolor{lila}{$\bullet$}) representing models trained from different random initialization. 
    Models within the green region (\textcolor{green}{$\bullet$}) cannot be distinguished at the 95\% significance level. 
    Notably, only two models from the Abdominal CT/ConvNeXt combination show a statistically significant performance difference---an effect that would likely remain undetected without formal statistical testing. 
    The dashed line (\textcolor{pink}{$\bullet$}) marks the model for which we performed a hyperparameter search. 
    Axes are scaled uniformly within each dataset (0.02 units per grid cell).}
    \label{fig:model_multiplicity}
\end{figure}

\noindent The true Rashomon set is the set of statistically indistinguishable \enquote{good} models. 
In practice, we face two constraints: model quality can only be assessed on finite samples from the underlying distribution (typically validation and test sets), and for complex model classes, an exhaustive characterization of the Rashomon set is infeasible---for example, in neural networks the size of the Rashomon set is tied to the number of local minima, which grows exponentially with the number of parameters \cite{auer1995exponentially}.
Thus, following prior research, we explore the Rashomon \textit{empirically} by varying the random seed used to initialize the weights of the last layer. 
This approach results in a set of models with substantial variation in their performance on the finite validation and test set; for example, accuracy differs by up to 16\% (see Breast Ultrasound/ResNet50 in \autoref{fig:model_multiplicity}).
While prior work, e.g., \cite{coston2021characterizing, hsu2022rashomon}, typically relies on \textit{ad hoc} performance thresholds (such as a fixed 1\% loss tolerance), we instead adopt the framework of \citet{paes2023inevitability}. 
We use the \gls{CP} interval \cite{clopper1934use}, an exact method for constructing confidence intervals for binomial error rates, and apply it to model accuracies (the inverse of error). 
Using the most accurate model among the 50 in the empirical Rashomon set as a reference $\epsilon_0$, we determine the smallest decrease in accuracy $\epsilon$ that leads to the rejection of the null hypothesis of equal true error rates, based on validation and test performance.
The width of this region depends on the Type-I error level ($\alpha$=0.05), the baseline error $\epsilon_0$ of the best empirical model, and the number of validation and test samples $n$.
We compute the rejection threshold $\epsilon$ numerically using bisection (see \Cref{subsec:testing-framework} for details and a complementary instance-level analysis). 

\autoref{fig:model_multiplicity} visualizes the \textit{indistinguishability region} between the reference performance $\epsilon_0$ and the rejection threshold $\epsilon$. 
All obtained models--- except for two in the Abdominal CT/ConvNeXt combination---cannot be distinguished at the 95\% significance level. 
In other words, while their validation and test accuracies may suggest varying quality, \textit{statistical testing provides no evidence that their performance on the underlying distribution is truly different}. 
Interestingly, the two models which \textit{are of inferior quality} are not extreme cases (compared to the other models in the empirical Rashomon set) and would have likely gone unnoticed without formal testing. 
Moreover, within the empirical Rashomon set, validation performance poorly predicts test performance: models that perform well on the validation set often underperform on the test set, and \textit{vice versa}. 
Notably, the initial model (dashed line \textcolor{pink}{$\bullet$} in \autoref{fig:model_multiplicity}) for which we performed the hyperparameter search is in no way \enquote{special} in terms of validation or test set performance relative to the other models. 
Better and worse models exist despite this being the model that has been explicitly optimized for. 

In summary, these findings imply that validation and test metrics alone cannot uniquely identify an \enquote{optimal} model. 
Crucially, even the initial model for which we perform a hyperparameter search is as good a draw as any other model from the set. 
Selecting the model with the highest validation score becomes an arbitrary decision among other equally valid alternatives. 
Consequently, the standard selection criteria---choosing the model with the best validation performance---is not only inadequate but potentially harmful to patients receiving inferior predictions. 

\subsection{Individual predictions are arbitrary under any single model}\label{subsec:arbitrariness}
So far, we have focused on the average performance metrics within the empirical Rashomon set. 
We now turn to model predictions to examine the \textit{arbitrariness of diagnostic outcomes when relying on a single model}.
We define \gls{APPA} to quantify prediction stability at the per-sample level, i.e., \textit{vis-a-vis} a patient.
\gls{APPA} measures the probability that two models, drawn uniformly at random from the empirical Rashomon set, assign the same diagnosis to a patient, normalizated by the number of models and classes (see \Cref{meth:appa} for a detailed derivation).
By definition, APPA ranges from 0 to 1 with APPA = 1 corresponding to maximal expected agreement between two models (high stability), APPA = 0 to maximal expected disagreement (low stability), and intermediate values indicate partial expected agreement. 

\begin{figure}[!h]
    \centering
    \includegraphics[width=1.\linewidth]{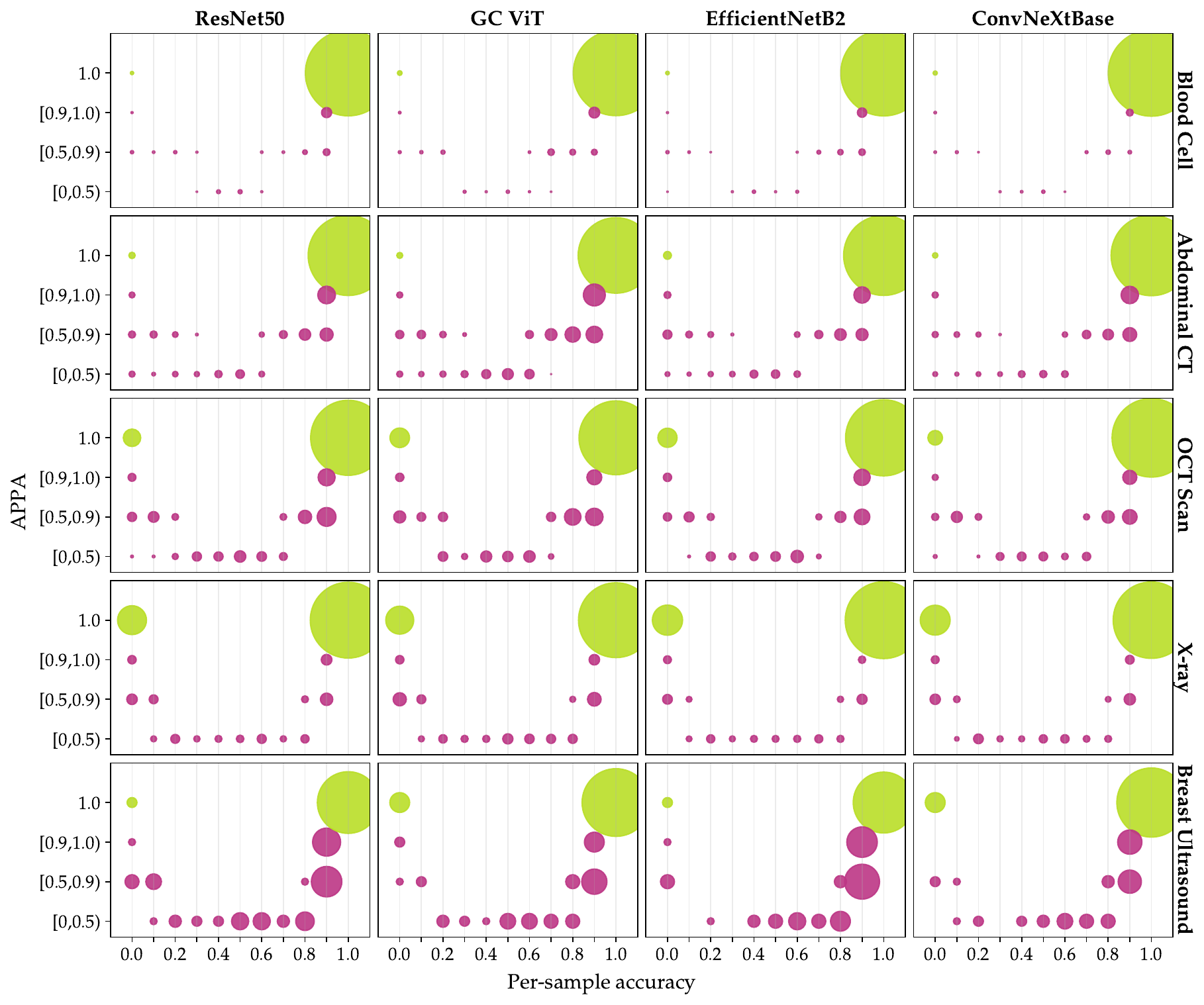}
    \caption{\textbf{Prediction stability as a function of accuracy.}
    For clarity, samples are binned by per-sample accuracy and APPA; point size reflects their relative frequency (normalized by dataset size). 
    Color encodes stability: (\textcolor{green}{$\bullet$}) marks stable samples (APPA = 1.0), whereas (\textcolor{pink}{$\bullet$}) correspond to unstable samples (APPA $<$ 1.0).
    Across datasets and architectures, high accuracy across the models in the empirical Rashomon set consistently exhibits lower APPA (top-right corner). 
    However, samples in the top-left corner are consistently misclassified, revealing systematic failure modes. 
    }
    \label{fig:APPA_and_acc}
\end{figure}

As illustrated by the pink points (\textcolor{pink}{$\bullet$}) in \autoref{fig:APPA_and_acc} arbitrary predictions occur across all datasets and architectures, though their extent varies considerably. 
For example, only 1.2\% of samples are affected for Blood Cell/ConvNeXt, whereas 50\% are affected for Breast Ultrasound/EfficientNetB2. 
Importantly, (dis)agreement is not uniformly distributed across the data but concentrated on \textit{specific} samples: those that are frequently predicted correctly (high accuracy) also exhibit high inter-model agreement (high APPA). 
This positive relationship is expected as both metrics depend on how models distribute their predictions across classes; intuitively, when multiple models predict the correct class, they necessarily agree more often. 
Although per-sample accuracy and agreement are correlated by definition, they capture distinct aspects of model behavior. 
\textit{Predictive multiplicity is not about whether a prediction is correct, but whether it could have been different under an equally valid model.}
Accuracy measures correctness relative to the ground truth, while predictive multiplicity reflects the consistency of predictions across models---regardless of correctness. 
Consequently, high \gls{APPA} is often associated with correct predictions but can also occur for incorrect ones, indicating systematic bias shared across models (see top-left regions in \autoref{fig:APPA_and_acc}).
In contrast, low \gls{APPA} implies that a prediction depends strongly on the specific model, i.e., a prediction which is arbitrary under any single model. 

The relationship between accuracy and predictive multiplicity extends to the model level. 
Intuitively, error provides the \enquote{space} for disagreement: a higher error rate creates more opportunities for conflicting predictions. 
Indeed, the \gls{GDE} \cite{jiang2021assessing} formalizes this connection: the expected disagreement rate between independently trained models (e.g., with different random seeds) approximately equals their error rate. 
More precisely, this equality holds for well-calibrated ensembles, a property that SGD-trained networks, such as ours, naturally exhibit \cite{lakshminarayanan2017simple}. 

In summary, predictive multiplicity poses a serious risk for high-stakes applications:
Relying on any single model exposes some patients to arbitrary outcomes, \textit{predictions not driven by meaningful patterns in the data but ultimately determined by a random seed}.
Yet, analyzing predictive multiplicity alongside accuracy offers a lens to analyze where and why models converge/diverge, offering both diagnostic insight (e.g., bias detection) and practical means (e.g., stability-based filtering) for building more reliable systems. 
Importantly, \textit{predictive multiplicity is not inherently negative---it can help identify bias and expose misclassifications that would otherwise go unnoticed}.

\subsection{Reducing Predictive Multiplicity Through Accuracy Maximization}\label{subsec:acc-and-mult}
Before moving on how to leverage predictive multiplicity in practice, we take a closer look at the relationship between model capacity, accuracy, and predictive multiplicity. 
Overparameterized networks---where the number of models exceeds the number of training samples---have been shown to generalize effectively, e.g., \citet{allen2019learning}.
According the the \gls{GDE} \cite{jiang2021assessing}, we expect networks with lower test error to exhibit less predictive multiplicity. 
However, \citet{black2022model} theoretically show that multiplicity is closely liked to variance. 
In particular, when higher accuracy is achieved by increasing model complexity (and thus variance), we should expect an increase in predictive multiplicity.
Intuitively, models with greater capacity can learn more intricate decision boundaries, which, in theory, allow for higher predictive multiplicity.
This relationship is further supported by their empirical findings \cite{black2021leave, black2021consistent}, which compares lower-complexity linear models to (highly) expressive deep neural networks. 
We build on those findings by examining predictive multiplicity in deep neural networks of different capacities. 
To operationalize capacity, we utilize EfficientNet variants, specifically EfficientNetB0 (5.3M parameters) and EfficientNetB4 (19.5M parameters). 
Both EfficientNet variants operate in the overparameterized regime for datasets consisting of 546 to 223,414 training samples. 

\autoref{tab:efficientnets} summarizes the population-level effects of increasing model capacity. 
For Blood Cell, Abdominal CT, and X-ray using a higher-capacity model (B4 instead of B0) results in little change in accuracy and APPA ($\leq 0.6\%$).
Overall utility remains stable, but note that the affected samples (last row in Table~\ref{tab:efficientnets}) differ across models. 
In contrast, for OCT Scan and Breast Ultrasound we see clear improvements---4.4\% improvement in accuracy and up to 12.8\% more stable samples. 
This pattern is also reflected in the proportion of affected samples---those whose stability or correctness changes between models. 

\begin{table}[hbt]
\caption{\textbf{Performance and stability across model capacities---when higher-capacity models achieve higher accuracy, stability increases (higher APPA); otherwise the aggregated effects are minimal.}
The first two rows report \textit{mean test accuracy} in \% ($\pm$ std) for EfficientNetB0 (relatively low capacity) and B4 (high capacity) across 50 models from the empirical Rashomon set.
Subsequent rows show the \textit{effects of increasing model capacity from B0 to B4} (in \%): the increase in mean accuracy ($\uparrow$ Acc), the change in mean \gls{APPA}, and the corresponding proportion of samples whose \gls{APPA} values differ between models (Affected).
For $\Delta$ APPA and Affected we report actual/binarized APPA values; for binarized APPA=1 when actual APPA=1, 0 otherwise.}
\centering
\begin{tabular}{lccccc}
\toprule
 & Blood Cell & Abdominal CT & OCT Scan & X-ray & Breast Ultrasound \\
\midrule
Accuracy B0 & 99.0 $\pm$ 0.1 & 95.2 $\pm$ 0.3 & 86.7 $\pm$ 1.3 & 83.0 $\pm$ 0.3 & 84.7 $\pm$ 3.7\\
Accuracy B4 & 99.1 $\pm$ 0.1 & 95.9 $\pm$ 0.3 & 91.1 $\pm$ 0.8 & 82.8 $\pm$ 0.2 & 89.1 $\pm$ 1.4\\
\midrule

$\uparrow$ Acc & 0.1 & 0.7 & 4.4 & -0.2 & 4.4 \\

$\Delta$ \gls{APPA} & -0.1 / -0.6 & -0.1 / -0.6 & 2.7 / 9.9 & 0.2 / 0.1 & 12.8 / 12.8\\
Affected & 3.9 / 1.9 & 16.0 / 5.4 & 25.9 / 14.3 & 13.4 / 5.2 & 66.0 / 30.8 \\

\bottomrule
\end{tabular}\label{tab:efficientnets}
\end{table}

\begin{figure}[htb]
    \centering
    \includegraphics[width=1\linewidth]{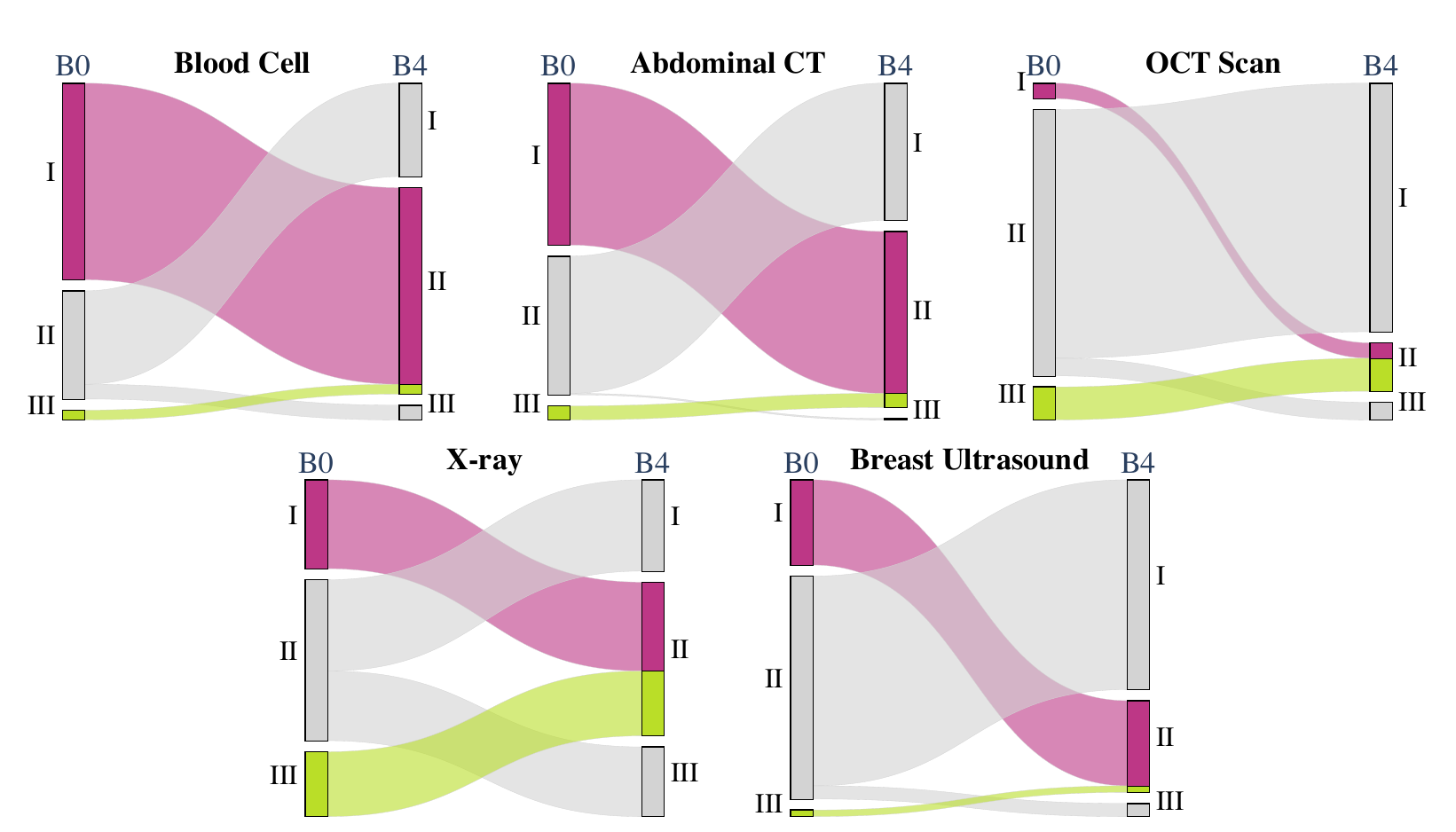}
    \caption{
    \textbf{Changes by increasing model capacity (only affected samples).}
    Samples are grouped as (I) correct-stable, (II) unstable, and (III) incorrect-stable.
    Bundles connect category transitions from EfficientNetB0 (left) to EfficientNetB4 (right), illustrating how samples shift between groups.
    Pink bundles mark the cost of model switching (previously stable-correct samples becoming unstable) while green bundles indicate newly detectable errors (previously consistently misclassified samples become unstable and thus identifiable). 
    For Blood Cell the effect is minimal (only 2.6\% of samples are affected), for Abdominal CT, OCT Scan, and Breast Ultrasound overall utility improves, for X-ray utility stays the same, but different samples are affected.
    }
    \label{fig:acc_and_multiplicity}
\end{figure}

To understand how the composition changes by increasing model capacity, we examine how the affected samples are distributed across categories in Figure~\ref{fig:acc_and_multiplicity}.
Two categories are of specific interest: 
First, previously stable samples that become unstable.
These represent the \textit{cost of model switching} (\textcolor{pink}{$\bullet$} in Figure~\ref{fig:acc_and_multiplicity}): cases that were previously handled reliably but are now predicted inconsistently depending on the specific, selected model. 
As we will see in \Cref{subsec:stability} we can detect these samples when considering more than one model as they are unstable across models.
Second, previously \textit{undetectable errors} (\textcolor{green}{$\bullet$}), samples that were consistently incorrect across models, but become inconsistent and thus detectable. 
We see a decrease in this fraction for the datasets which achieve higher accuracy with the higher capacity model (Abdominal CT and OCT Scan). 
Reducing the size of this category is desirable, as these samples are consistently misclassified across models and remain unidentifiable even when multiple models are considered.

In summary, our results suggest that in already overparameterized models capacity alone does not determine predictive multiplicity. 
When increasing capacity does not improve accuracy, we observe only marginal changes in predictive multiplicity (affecting only a small fraction of samples); however, when higher capacity leads to improved accuracy, predictive multiplicity consistently decreases. 
This indicates that the relationship between capacity and multiplicity is mediated by accuracy: greater expressiveness can either introduce minor additional variability when performance remains stagnant or, when it enhances generalization, substantially reduce predictive multiplicity.
Yet, awareness and systematic monitoring of such shifts in the composition of affected samples---and of which individuals/groups are impacted---remain essential, particular in high-stakes domains like healthcare, where collective performance gains must be carefully balanced against individual rights.  

\subsection{Prediction reliability requires more than one model}\label{subsec:stability}
\begin{figure}[htbp]
    \centering
    \includegraphics[width=1\linewidth]{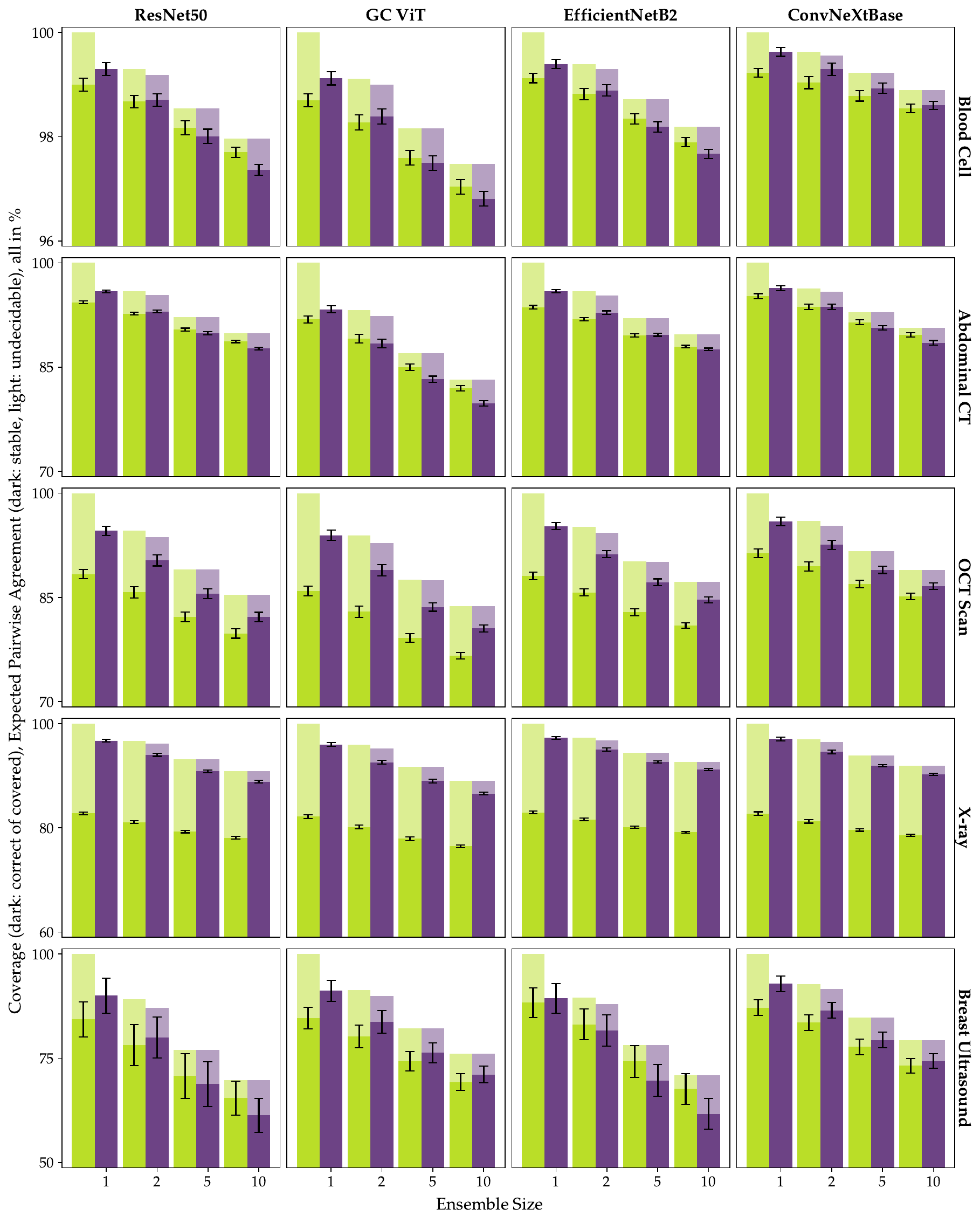}
    \caption{
    \textbf{Ensembles substantially reduce predictive multiplicity.}
    Coverage rate (\textcolor{green}{$\bullet$}) and expected pairwise agreement (\textcolor{lila}{$\bullet$}) of the test set across single models and ensembles of sizes 2, 5, and 10. 
    For \textit{coverage}, the darker segment indicates the correctly covered samples. 
    For \textit{expected pairwise agreement}, the darker segment represents stable samples, while the lighter segment corresponds to samples for which no judgment can be made due to missing coverage from the other ensemble.
    Error bars show standard deviation.
    Note the X-ray dataset: ensembles reduce predictive multiplicity but retain consistently incorrect predictions, reflecting shared systematic biases.
    }
    \label{fig:stability}
\end{figure}

Using more than one model allows the detection of predictions that would be arbitrary under a single model. 
Further, we can abstain from predicting when there is insufficient consensus \cite{black2022selective}, and flag ambiguous and potentially harmful predictions.
The capacity to abstain does not come without costs: while it improves reliability and robustness, it requires multiple models and may reduce coverage, as not all samples receive predictions.
To evaluate the effectiveness, we compare \textit{coverage rates}, \textit{correctness} (across covered samples), and \textit{predictive stability} across single models and ensembles consisting of two, five, and ten models. 
To assess predictive stability, we compute the \textit{expected pairwise agreement} between two distinct models or ensembles of equal size (see \Cref{meth:ensembling} for more details).
Intuitively, this metric captures how often two equally plausible models/ ensembles agree, complementing existing measures of disagreement and discrepancy \cite{black2022model, marx2020predictive, d2022underspecification}.
For the ensembles, we apply a conservative decision rule: a prediction is made \textit{only if all constituent models agree}; otherwise, the ensemble abstains.
Note that we can assess agreement between ensembles only on samples where both ensembles made a prediction; we display the fraction of samples that cannot be evaluated because they are not predicted by the alternative ensemble in (\textcolor{lila}{$\bullet$}, low opacity in \autoref{fig:stability}). 

As shown in \autoref{fig:stability}, using ensembles reduces predictive multiplicity across all datasets and model architectures. 
The main exceptions are the Breast Ultrasound dataset with ResNet50 (for all ensemble sizes) and EfficientNetB2 (for ensembles of size five and ten) where stability decreases and Blood Cell for which stability is already very high ($>$ 99\%) and the room for improvement is very small (see \autoref{app-tab:stability} and \autoref{fig:app-ensembling} in the Appendix for more details).
The reported stability values for ensembles (in comparison to single models) should be interpreted as conservative estimates, since samples for which another ensemble of the same size would abstain from making a prediction are excluded; for these cases, stability cannot be assessed.
With respect to accuracy, and in line with prior work, ensembles improve predictive performance (see, for example, \citet{dietterich2000ensemble} for an overview).
This improvement is reflected in an increased proportion of correct predictions among the covered samples, as indicated by the expansion of the darker regions within the green bars in \autoref{fig:stability} (see also \autoref{fig:app-ensembling} in the Appendix for a sample-wise analysis).
This demonstrates that ensembles not only reduce predictive multiplicity (compared to an alternative, equally plausible predictor), but also concentrate their predictions on cases where they are more likely to be correct.
Note the X-ray dataset as an interesting case: while ensembles reduce predictive multiplicity, some samples are consistently incorrect classified, suggesting that ensemble members converge to the same systematic biases.
This underscores that ensemble agreement should not be conflated with correctness and motivates the need for complementary bias-aware or data-centric interventions.

To sum up, using ensembles instead of single models improves both the stability of predictions (with respect to alternative models or ensembles) and predictive accuracy. 
This improvement comes at the cost of reduced coverage. 
Notably, ensembles improve correctness only for samples that are ambiguous across ensemble members (i.e., those that suffer from predictive multiplicity); they do not aid in detecting samples that are misclassified consistently.

\section{Discussion}\label{sec:discussion}
We presented a comprehensive evaluation of the empirical Rashomon set in the medical domain and show that the existence of multiple equally valid models challenges conventional practices of model selection and deployment.
While grounded in healthcare, our findings likely extend to other high-stakes domains facing similar constraints of uncertainty, limited data, and ethical responsibility. 
Traditionally, studies of the Rashomon effect consider only models with small performance differences as statistically indistinguishable.
Using the framework of \citet{paes2023inevitability}, we find that even models with substantial differences can be statistically indistinguishable, underscoring the limits of conventional performance-based comparisons. 
This aligns with previous work \cite{jordan2024variance}, which demonstrates that apparent performance variance across finite test sets often reflects finite-sample noise rather than true generalization differences---a problem potentially amplified in medicine, where datasets are increasingly large overall but still exhibit substantial representation gaps for rare or heterogeneous conditions.
Consequently, sampling variability may distort the empirical Rashomon set by overfitting idiosyncrasies of the finite test set. 
These findings highlight important directions for future research and call for rethinking statistical and methodological practices in model evaluation (see also \autoref{app:related-work}).

The existence of multiple models with equal quality renders the selection of a single, supposedly \enquote{best}-performing model effectively arbitrary, undermining the justification for the choice of model and its predictions. 
Relying on one such model exposes some patients to effectively arbitrary and potentially harmful outcomes---those whose predictions would differ under another, equally valid model. 
\citet{creel2022algorithmic} argue that isolated arbitrary decisions are not inherently morally problematic, except when other rights make non-arbitrariness normatively relevant. 
In medicine, such rights arguably exist: patients are entitled to informed consent\footnote{
While the form and legal status of informed consent vary across cultural and regulatory contexts, its underlying commitment to respecting patient self-determination is widely recognized \cite{angell1988ethical}.
Arbitrariness in model selection may compromise this principle, as patients cannot meaningfully consent to decisions whose rationale varies unpredictably between equally valid models.
} and to consistent, evidence-based, and non-arbitrary treatment \cite{olejarczyk2024patient, varkey2021principles}. 
While we identified technical sources of arbitrariness, questions such as \textit{to what extent should patients have a right to non-arbitrary, consistent treatment?} and \textit{how should informed consent be interpreted under model multiplicity?} require further ethical and legal analysis. 
In how far existing regulatory frameworks---such as the EU Artificial Intelligence Act and related data governance provisions---already capture these concerns warrants domain expert examination. 
If left unaddressed, model multiplicity, and particularly predictive multiplicity, may complicate the ethical and legal justification of AI-assisted decision-making in healthcare.

Contrary to \citet{black2022model}, we find that higher-capacity models can enhance both accuracy and predictive stability.
We agree, however, that \enquote{accuracy is not an antidote to multiplicity, and model selection cannot simply be reduced to accuracy-maximization} \cite{black2022model}---even more, accuracy-maximization is an insufficient criterion for model selection. 
However, in real-world applications, where overparameterized models are the norm, the number of trainable parameters may, in fact, have limited relevance for predictive multiplicity.

Samples with higher expected accuracy are predicted more consistently across models, suggesting that predictive multiplicity is not inherently detrimental but can be leveraged beyond the single-model paradigm. 
Using ensembles and predicting only on consistent samples improves both accuracy and stability. 
An ensemble with selective abstention---deferring unstable cases to human review---eliminates measurable multiplicity and aligns predictive confidence with clinical accountability.
Our consensus-based approach, which requires unanimous model agreement, is intentionally simple.
More sophisticated methods, such as the statistical consistency test by \citet{black2022selective}, may be better suited for deployment. 
The optimal agreement criterion and ensemble size depends on the application domain, balancing computational cost and desired confidence. 

Our work is not without limitations. 
We do not fully characterize the broader impact of predictive multiplicity on medical diagnosis tasks, and we focus exclusively on classification problems thereby omitting the full diversity of clinical scenarios and modeling paradigms. 
In line with this, we do not assess how predictive multiplicity influences downstream clinical decision-making or patient outcomes---an important area for future research. 
Our aim is to demonstrate the implications of predictive multiplicity in medical application domains and to motivate further investigation into mitigating the risks that model multiplicity poses to the adoption of machine learning in high-stakes domains.

\section{Conclusion}\label{conc}
In this study we show that predictive multiplicity is both pervasive and consequential in medical AI. 
Small, seemingly inconsequential training variations can lead to different predictions for individual patients despite statistically indistinguishably overall performance of the models. 
This finding challenges the widespread assumption that a single \enquote{best} model can reliably guide decisions in high-stakes (clinical) contexts.
Our results reveal that multiplicity is not a rare anomaly but a fundamental property of modern predictive modeling.
Recognizing it as a structural property of the learning process---rather than a nuisance---calls for a rethinking of how models are evaluated, selected, and deployed in high-stakes settings. 
Ultimately, predictive multiplicity matters most where it is least tolerable, at the point of care, where treatment decisions depend on individual predictions.
By acknowledging and characterizing predictive multiplicity, we seek to catalyze the development of diagnostic machine learning systems that are not only accurate, but also robust, equitable, and trustworthy.

\section{Methods}\label{sec:methods}
In the following sections, we provide a detailed description of the datasets, model architectures, and training procedures. 
We then present our numerical estimation of the Rashomon parameters, derive the \gls{APPA} metric, and outline our methodology for evaluating ensembles.

\subsection{Data availability statement}
We used the CheXpert chest X-ray dataset under special authorization from the Stanford ML Group (https://stanfordmlgroup.github.io/competitions/chexpert/). 
All other datasets used in this study are publicly available via MedMNIST (https://medmnist.com). 
Code to reproduce the experiments will be made publicly available upon acceptance of the manuscript.

\subsection{Datasets, model architectures and model training}
The five medical imaging datasets span diverse modalities and classification tasks: Abdominal CT \cite{bilic2023liver}, Breast Ultrasound \cite{al2020dataset}, Blood Cell \cite{acevedo2020dataset}, OCT Scan \cite{kermany2018identifying}, and X-Ray \cite{irvin2019chexpert}. 
See Table~\ref{app-tab:sample-counts} for an overview of dataset properties. 
All experiments rely on the official training, validation, and test splits to ensure comparability with prior work.
For X-ray, we adopted the protocol of \citet{irvin2019chexpert}, training multi-label binary classifiers across all 14 classes and evaluating performance on the five pathologies with annotated validation/test labels: Atelectasis, Cardiomegaly, Consolidation, Edema, and Pleural Effusion. 
Each image was treated as an independent sample. 
Uncertain labels were converted to binary: mapped to positive for Atelectasis, Edema, and Pleural Effusion, and to negative for all other classes.
For all others, see \citet{medmnistv2}.
X-ray images were resized to 1$\times$224$\times$224. 
All other datasets followed the MedMNISTv2 preprocessing \cite{medmnistv2}, resulting in C$\times$224$\times$224 inputs (C=3 for Blood Cell; C=1 for the others). 
Single-channel inputs were triplicated to match the input dimensionality expected by ImageNet-pretrained models.

\begin{table}[ht]
\centering
\caption{Number of training, validation, test samples, and classes for each dataset.
*The training set contains 14 labels, with five designated as target labels (details provided in the text.)}
\label{app-tab:sample-counts}
\begin{tabular}{lcccc}
\toprule
Dataset & \# Training & \# Validation & \# Test & \# Labels \\
\midrule
Abdominal CT  & 34,561 & 2,392  & 8,825  & 11 \\
Blood Cell   & 11,959 & 1,712  & 3,421  & 8 \\
Breast Ultras.  & 546 & 78     & 156    & 2 \\
OCT Scan     & 97,477 & 10,832 & 1,000  & 4 \\
X-ray     & 223,414 & 234 & 500  & 14/5* \\
\bottomrule
\end{tabular}
\end{table}

We evaluated four model architectures ConvNeXt \cite{liu2022convnet}, EfficientNet\cite{tan2019efficientnet} (variants B0 and B4 in subsection~\ref{subsec:acc-and-mult}; B2 for all others), GC ViT \cite{hatamizadeh2023global}, and ResNet50 \cite{he2016deep}, all pretrained on ImageNet \cite{deng2009imagenet}.
To differ random weight initialization, we replaced the final classification layer with a randomly initialized dense layer matching the number of classes in the respective dataset, using a Glorot uniform initializer \cite{glorot2010understanding}.
All training was conducted under deterministic conditions with fixed random seeds. 
Within each dataset/architecture combination, variation across models (i.e., the Rashomon set exploration) was induced solely by varying the random seed, which determined the initial weights of the final classification layer via the Glorot uniform initializer. 
All other factors were fixed to ensure reproducibility. 
Models were trained with a batch size of 64 using the AdamW optimizer \cite{loshchilov2017decoupled}, with exponential decay rates of 0.9 and 0.999 for the first and second-moment estimates, respectively. 
To select the initial learning rate, we performed a sweep over three learning rates, namely {0.01, 0.001, and 0.0001}, with a fixed random seed (seed = 0); the best-performing learning rate was used in all subsequent experiments without further tuning. 
We employed a cosine decay learning rate schedule \cite{loshchilov2016sgdr}.
For the X-ray dataset, the decay steps equaled the total number of training iterations (epochs $\times$ steps per epoch), while for all other datasets, the decay steps matched the number of epochs.
We trained for a fixed number of epochs without early stopping, with the number of epochs depending on the dataset: 15 epochs for Breast Ultrasound, five epochs for Blood Cell, OCT Scan, and Abdominal CT, and one for X-ray.
We used sparse categorical cross-entropy for all other datasets. 
Classification accuracy served as the primary performance metric.
Table~\ref{app-tab:performance} reports both the performance of the initial model and the mean performance across the 50 models from the empirical Rashomon set.
Across datasets, our models perform competitively and often surpass reported benchmarks.
All models and training procedures were implemented using Keras 3.8.
We used different GPUs for different dataset/architecture combinations; however, all model instances within one empirical Rashomon set, i.e. dataset/architecture combination, were trained on the same GPU to ensure consistency.

\begin{table}[thb]
\caption{
For each dataset (columns) and model architecture (rows), we report the initial model's test performance, along with the mean test accuracy across 50 models from the empirical Rashomon set, in parentheses. 
The best-performing initial model for each dataset is highlighted in bold.
Benchmark results are from \citet{irvin2019chexpert} for X-ray (highest ROC-AUC across five uncertainty methods) and from \citet{medmnistv2} for all other datasets (highest accuracy across seven architectures).}

\centering
\begin{tabular}{lccccccc}
\toprule
 & Benchmark & ResNet50 & GC ViT & EfficientNetB2 & ConvNeXt \\
\midrule
Abdominal CT        & 92.0      & 94.6 (94.4)   & 91.2 (91.6)   & 93.5 (93.6)   & \textbf{95.3} (95.2) \\
Blood Cell          & 96.6      & 98.9 (99.0)   & 98.7 (98.7)   & 99.0 (99.1)   & \textbf{99.2} (99.2) \\
Breast Ultrasound   & 86.8      & 80.8 (83.9)   & 84.6 (84.6)   & \textbf{89.7} (88.5)  & 88.5 (87.1) \\
OCT Scan            & 77.6      & 87.4 (88.3)   & 87.4 (86.0)   & 86.4 (88.1)   & \textbf{92.5} (91.4) \\
X-ray               & 89.5 & 88.9 (88.6)   & 87.2 (87.6)  & \textbf{89.8} (89.9)  & 89.7 (89.7) \\
\bottomrule
\end{tabular}\label{app-tab:performance}
\end{table}

\subsection{Numerical Rashomon Parameter Estimation}
\label{subsec:testing-framework}
\label{app:testing-framework}
To identify the range of statistically indistinguishable models within the empirical Rashomon set, we determine the smallest error rate $\epsilon$ for which the \gls{CP} confidence intervals of two models' empirical errors no longer overlap. 
This point marks the boundary at which we can reject the null hypothesis that both models have equal true error rates at significance level $\alpha$ (we choose $\alpha=0.05$).

\begin{table}[ht]
\caption{The Rashomon parameter $\epsilon$ and the reference model $\epsilon_0$.}

\centering
\begin{tabular}{llrrrr}
\toprule
\multirow{2}{*}{Dataset} & \multirow{2}{*}{Model} 
    & \multicolumn{2}{c}{Validation} 
    & \multicolumn{2}{c}{Test} \\
& & $\epsilon_0$ & $\epsilon$ & $\epsilon_0$ & $\epsilon$ \\
\midrule
Breast & ResNet & 5.1 & 21.8 & 9.6 & 21.8\\
& GC ViT & 5.1 & 21.8 & 12.2 & 25.0\\
& EfficientNetB2 & 2.6 & 16.7 & 7.1 & 17.9\\
& ConvNeXt & 5.1 & 21.8 & 10.9 & 23.7\\

\midrule
Blood Cell & ResNet & 0.7 & 1.8 & 0.7 & 1.4\\
& GC ViT & 0.9 & 2.1 & 1.1 & 1.9\\
& EfficientNetB2 & 0.6 & 1.8 & 0.7 & 1.5\\
& ConvNeXt & 0.5 & 1.6 & 0.6 & 1.3\\

\midrule
Abdominal CT & ResNet & 0.6 & 1.4 & 5.2 & 6.2\\
& GCViT & 1.2 & 2.3 & 7.0 & 8.2\\
& EfficientNetB2 & 0.6 & 1.4 & 5.8 & 6.9\\
& ConvNeXt & 0.4 & 1.2 & 4.0 & 4.9\\

\midrule
OCT Scan & ResNet & 2.0 & 2.6 & 9.1 & 13.1\\
& GCViT & 2.1 & 2.7 & 11.9 & 16.4\\
& EfficientNetB2 & 1.9 & 2.5 & 10.4 & 14.6\\
& ConvNeXt & 1.6 & 2.2 & 6.4 & 9.9\\

\midrule
X-ray & ResNet50 & 17.9 & 22.6 & 16.8 & 19.9 \\
& GCViT & 18.5 & 23.2 & 17.4 & 20.5 \\
& EfficientNetB2 & 17.9 & 22.6 & 16.4 & 19.4 \\
& ConvNeXt & 17.8 & 22.5 & 16.9 & 20.0 \\

\bottomrule
\end{tabular}
\label{tab:epsilon}
\end{table}

We compute $\epsilon$ numerically using a bisection search.
Given the number of samples in the respective (finite) dataset $n$ and $\epsilon_0$, which we define as the lowest obtained error rate among the 50 models in the empirical Rashomon set, we first compute its upper \gls{CP} bound $UB(\alpha, n\epsilon_0, n)$ (see Table~\ref{app-tab:sample-counts} for the number of samples per dataset and split; see Table~\ref{tab:epsilon} for $\epsilon_0$ and $\epsilon$ values).
We then search for the smallest $\epsilon$, such that $LB(\alpha, n\epsilon, n) = UB(\alpha, n\epsilon_0, n)$, where $LB$ and $UB$ are the lower and upper \gls{CP} confidence limits respectively.
The bisection procedure iterates until convergence (tolerance $<10^{-8}$) or after 10,000 steps. 
We adopt an exact binomial model by rounding $n\epsilon$ to the nearest integer, ensuring that the \gls{CP} bounds are computed from valid discrete sample counts. 
Note that overlapping confidence intervals do not constitute a formal statistical test for evaluating the equality of means (e.g., \citet{schenker2001judging}); the approach is conservative and may miss small but statistically significant differences. 
However, its simplicity and graphical interpretability make it appealing for practitioners (see \citet{paes2023inevitability} for more details and alternatives). 

The Rashomon parameter $\epsilon$ is determined by the baseline error $\epsilon_0$, the dataset size $n$, and the confidence level ($1-\alpha$).
Intuitively, a smaller $\alpha$ (i.e., higher confidence) leads to wider \gls{CP} intervals, which in turn increase the indistinguishability region. 
A larger $\epsilon_0$ (worse baseline performance) also enlarges the region, since performance differences must be larger before they become statistically meaningful. Finally, increasing $n$ narrows the region, because more data reduces statistical uncertainty and makes small differences detectable.

Model selection is often guided by \textit{marginal} error rates, i.e., a model's overall accuracy, which is well-captured by the \gls{CP} confidence intervals presented in the main body. 
McNemar's test \cite{mcnemar1947note} complements this perspective by assessing whether two models differ on the same instances: rather than comparing average error levels, it evaluates whether two models disagree on the \emph{same} instances, thereby revealing differences that marginal errors may obscure.

We first apply McNemar’s test to the \textit{first} model---the one selected via hyperparameter search---and assess how many of the 49 alternative models in the empirical Rashomon set cannot be distinguished from it at the 95\% significance level.
\autoref{app-tab:mcnemar} reports the proportion of alternative models that are statistically indistinguishable on the validation and test sets, i.e., the proportion for which the null hypothesis cannot be rejected ($p > 0.05$).
These results corroborate our conclusions in \autoref{subsec:rashomon-set}: the first model is not exceptional.
Across datasets, 85-100\% of models are statistically indistinguishable from it on the validation set, and the majority remain so on the test set (besides the Abdominal CT/ GCViT and OCT Scan/ EfficientNetB2).
For the datasets Abdominal CT and OCT Scan, we see more variability, especially w.r.t. the test set, suggesting split instability. 
Overall, according to McNemar's test, the first model appears far from unique, reinforcing that model selection is highly underdetermined.

We next apply the same procedure using the \textit{best} model among the 50 candidates as the reference model. 
Relative to the \textit{first} model, the proportion of statistically indistinguishable models becomes smaller, slightly for Blood Cell, Breast Ultrasound and X-ray, but more substantially for Abdominal CT and OCT Scan.
For example, on Abdominal CT with ConvNeXtBase, only 16.3\% of models are indistinguishable from the best model on the validation split, and 12.2\% on the test split.

In summary, while varying model initialization results in models that perform better according to McNemar's test, even the \enquote{best} model is far from unique: in every combination, there exist other models that cannot be statistically distinguished from it, indicating that it does not represent a single \enquote{optimal} solution.

\begin{table}[t]
\centering
\caption{\textbf{Percentage of models in the Rashomon set under McNemar’s test for each architecture/dataset pair.} 
For each entry, the left value corresponds to the validation split, and the right value corresponds to the test split. 
The upper block uses the first model as a reference, while the lower block uses the best model.}
\begin{tabular}{lcccc}
\toprule
\textbf{First model} & ResNet50 & GCViT & EfficientNetB2 & ConvNeXtBase \\
\midrule
Blood Cell     & 100.0 / 95.9 & 98.0 / 100.0 & 91.8 / 100.0 & 100.0 / 95.9 \\
Abdominal CT    & 61.2 / 85.7 & 95.9 / 34.7 & 100.0 / 89.8 & 85.7 / 67.3 \\
OCT Scan      & 100.0 / 63.3 & 100.0 / 55.1 & 83.7 / 40.8 & 95.9 / 51.0 \\
X-ray  & 100.0 / 98.0 & 100.0 / 100.0 & 100.0 / 79.6 & 98.0 / 93.9 \\
Breast Ultrasound   & 83.7 / 75.5 & 100.0 / 100.0 & 100.0 / 98.0 & 100.0 / 100.0 \\
\midrule
\textbf{Best model} & ResNet50 & GCViT & EfficientNetB2 & ConvNeXtBase \\
\midrule
Blood Cell    & 79.6 / 38.8 & 93.9 / 91.8 & 87.8 / 98.0 & 95.9 / 81.6 \\
Abdominal CT    & 61.2 / 46.9 & 51.0 / 16.3 & 83.7 / 26.5 & 16.3 / 12.2 \\
OCT Scan      & 85.7 / 18.4 & 71.4 / 34.7 & 83.7 / 36.7 & 63.3 / 16.3 \\
X-ray  & 91.8 / 85.7 & 93.9 / 89.8 & 95.9 / 49.0 & 91.8 / 93.9 \\
Breast Ultrasound   & 63.3 / 40.8 & 87.8 / 85.7 & 83.7 / 59.2 & 100.0 / 91.8 \\
\bottomrule
\end{tabular}

\label{app-tab:mcnemar}
\end{table}

\subsection{Adjusted Pairwise Prediction Agreement}
\label{meth:appa}
We assess per-sample prediction stability using \glsentrylong{APPA}, which quantifies the probability that two models, sampled uniformly at random from the empirical Rashomon set produce the same predictions.

Formally, let $M > 1$ denote the number of models, each predicting a class $c \in \{1,...,K\}$ for the same sample $x$.
For a given input $x$, let $n_c(x)$ represent the number of models that predict class $c$. 
The unnormalized pairwise prediction agreement is defined as
\begin{equation}
    \operatorname{PPA}(x) = \frac{\sum_{c=1}^K n_c(x)(n_c(x)-1)}{M(M-1)}.
\end{equation}
As the minimal attainable pairwise agreement depends on both the number of models $M$ and classes $K$ (intuitively, with fewer classes than models, some models must necessarily coincide in their predictions), we normalize by the achievable minimum $\operatorname{PPA_{min}}(M,K)$.
This minimum corresponds to model predictions being distributed as uniformly as possible across classes; with $q, r = M \bmod K$, $r$ classes receive $q+1$ predictions, and $(K-r)$ classes receive $q$ predictions, resulting in 
\begin{equation}
    \operatorname{PPA_{min}}(M,K)=\frac{(K-r)q(q-1)+r(q+1)q}{M(M-1)}.
\end{equation}
We then compute $\operatorname{APPA(x)}$ as
\begin{equation}
    \operatorname{APPA}(x)= \frac{\operatorname{PPA}(x) - \operatorname{PPA_{min}}(M,K)}{1 - \operatorname{PPA_{min}}(M,K)},
\end{equation}
which captures the excess agreement beyond what is expected from maximally uniform predictions, enabling comparison across tasks with differing numbers of models or classes. 
$\operatorname{PPA_{min}}$ is a reasonable default in the absence of prior knowledge about the distribution.
When information about the true distribution is available (such as the prevalence), using this prior can be a more appropriate choice.

\subsection{Ensembling predictions}\label{meth:ensembling}
We evaluate the effectiveness of ensembles to reduce predictive multiplicity by comparing prediction stability and coverage rates pairwise between single models and ensembles of size two, five, and ten.
Prediction stability is measured as the expected pairwise agreement between two distinct models (or ensembles of equal size) on the test set and averaged over 100 repetitions.
To avoid zero-inflation, we draw model or ensemble pairs without replacement from the empirical Rashomon set.
For ensembles, we apply a conservative decision rule: a prediction is made only if all constituent models agree; otherwise, the ensemble abstains.

\begin{table}[ht]
\centering
\caption{\textbf{Percentage of stable predictions are higher for ensembles than for single models, except for ResNet50 and EfficientNetB2 on the Breast Ultrasound dataset.}
We report the mean percentage of samples with stable predictions for a single model and an ensemble of size ten models over 100 repetitions.
For a single model, stability is computed over all test samples (full coverage) by comparing predictions to those of another single model.
For an ensemble, stability is computed only on samples for which all ensemble members agree and is evaluated by comparison with an ensemble of the same size.
Note, reported ensemble percentages are conservative estimates, as samples on which another ensemble of the same size would abstain from prediction are excluded; for these samples, no stability assessment can be made.
}
\label{app-tab:stability}
\begin{tabular}{lcccc}
\toprule
& ResNet50 & GCViT & EfficientNetB2 & ConvNeXtBase \\
\midrule
Blood Cell &99.30 / 99.39 &99.12 / 99.32 &99.39 / 99.46 &99.63 / 99.70  \\
Abdominal CT &95.89 / 97.54 &93.31 / 95.88 &95.91 / 97.60 &96.35 / 97.65 \\
OCT Scan &94.59 / 96.24 &93.95 / 96.19 &95.28 / 97.04 &95.93 / 97.42  \\
X-ray &96.72 / 97.79 &95.97 / 97.24 &97.26 / 98.46 &97.05 / 98.25  \\
Breast Ultrasound &89.99 / 87.85 &91.17 / 93.48 &89.40 / 86.97 &92.86 / 93.73  \\
\bottomrule
\end{tabular}
\end{table}
\begin{figure}[htbp]
    \centering
    \includegraphics[width=0.9\linewidth]{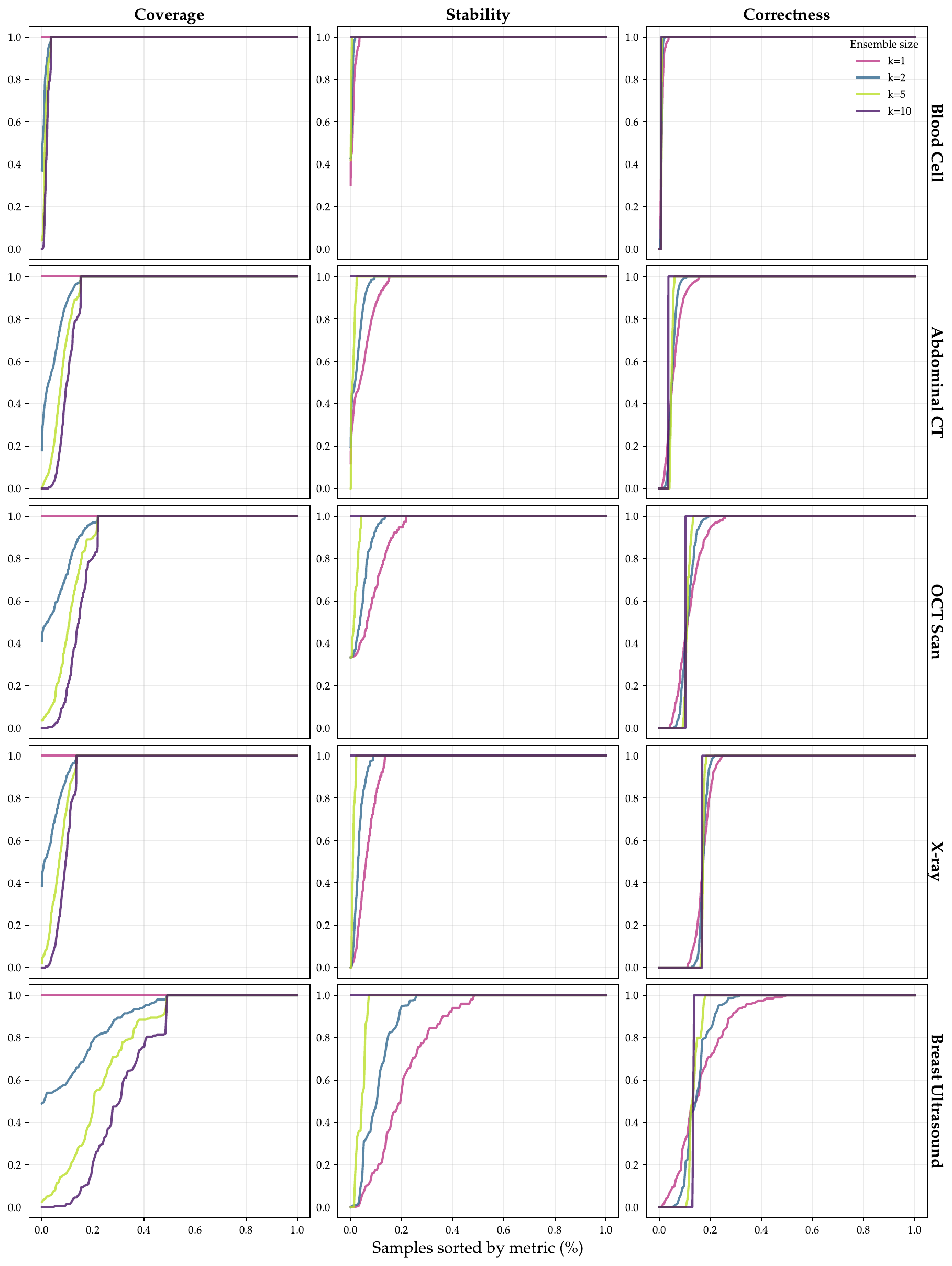}
    \caption{
    \textbf{Per-sample coverage, stability, and correctness as a function of ensemble size across datasets, exemplary for ResNet50.}
    For each dataset and ensemble size ($k \in \lbrace 1, 2, 5, 10 \rbrace$), we compute per-sample metrics over random disjoint single models or ensembles (of equal size) and visualize their empirical distributions by sorting samples according to the respective metric value. 
    The x-axis denotes the fraction of samples, and the y-axis shows the corresponding metric value. 
    \textit{Coverage} measures the probability that a sample is predicted by the ensemble, \textit{stability} (APPA) captures the agreement of predictions across sampled ensembles, and \textit{correctness} measures the probability that a prediction is correct given that a prediction is made. 
    Increasing ensemble size consistently reduces coverage while improving stability across datasets, with dataset-specific saturation behavior. 
    Correctness exhibits a heterogeneous response to ensembling, revealing distinct regimes of samples with limited versus substantial gains from larger ensembles.
    }
    \label{fig:app-ensembling}
\end{figure}

\autoref{fig:app-ensembling} shows per-sample distributions of coverage, stability (measured by \gls{APPA}), and correctness for increasing ensemble sizes across all datasets (for ResNet50). 
For each dataset and ensemble size $k \in \{1,2,5,10\}$, predictions are obtained from multiple random disjoint ensembles, and per-sample metrics are computed by aggregating over 100 ensemble draws. 
To enable distributional comparison, samples are sorted for each metric and ensemble size, and plotted as a function of the fraction of samples sorted by the respective metric.
\textit{Coverage} quantifies the probability that a sample receives a prediction from the ensemble. 
As expected, increasing ensemble size leads to lower coverage, indicating that larger ensembles are more conservative in issuing predictions. 
This is reflected by \textit{stability}, measured via \gls{APPA}, which captures the consistency of predictions across ensembles.  
As ensemble size increases, stability improves for a growing fraction of samples in all datasets.
The increase in stability with more ensemble members is little for the Blood Cell dataset which exhibits uniformly high stability even for a single model, indicating that predictions are already consistent and that additional ensemble members provide only marginal gains. 
\textit{Correctness}, defined as the probability of being correct conditional on making a prediction, reveals a more nuanced behavior. 
While increasing ensemble size generally improves correctness, the effect is heterogeneous across samples. 
In particular, the sorted correctness curves exhibit a change in curvature.
For lower-quantile samples, correctness improves slowly with ensemble size, indicating errors that are largely irreducible and likely dominated by systematic bias or intrinsic ambiguity. 
In contrast, higher-quantile samples benefit substantially from ensembling, with correctness increasing rapidly as ensemble size grows.

Taken together, these results show that increasing ensemble size trades coverage for stability and selectively improves correctness. While ensembles enhance predictive consistency and accuracy for a substantial subset of samples, they also expose a class of samples for which errors remain largely irreducible, highlighting the importance of per-sample analysis when assessing ensemble behavior.

\appendix
\section{Related Literature}\label{app:related-work}
In the following we provide a more detailed discussion of related literature that complements the brief overview presented in the main text. 
The phenomenon of model multiplicity was first described by \citet{breiman2001statistical} as the \textit{Rashomon Effect}.
They observed that small perturbations in the training set for decision trees and different weight initializations for small neural networks can lead to different solutions while having approximately equal error rates. 
More recent work showed that model multiplicity is ubiquitous in modern machine learning and a key obstacle to reliable training models that behave as expected in deployment \cite{d2022underspecification}.
The existence of multiple equally performing models is particularly relevant w.r.t.\@ their effect and consequences in the real world.
Several works highlight the opportunities that model multiplicity offers \cite{rudin2024position}, like the selection of fairer \cite{dutta2020there, wick2019unlocking}, more interpretable \cite{chen2018interpretable}, or more robust models \cite{d2022underspecification} without impairing predictive performance. 
Challenges arising from model multiplicity are among others the inconsistency of explanations \cite{hancox2020robustness, pawelczyk2020counterfactual}, the risk of fair-washing explanations \cite{anders2020fairwashing} or fairness metrics \cite{black2024d}, and predictive multiplicity \cite{marx2020predictive}---the main focus of this paper.

Despite the relevance of predictive multiplicity in the medical domain, systematic investigations into its risks and mitigation remains limited. 
To the best of our knowledge previous work in the medical domain has leveraged model multiplicity for trustworthy explanations \cite{kobylinska2024exploration}, explored the role of underspecification in model robustness \cite{d2022underspecification} and examined representational variability \cite{mehrer2020individual} (both works highlight the cause for predictive multiplicity without addressing how such instability/ variability affects predictions), and proposed bootstrapping as a remedy to predictive multiplicity \cite{riley2023clinical, riley2023stability}.

\textit{Bootstrapping} trains ensemble members on different resampled versions of the training data to induce diversity. 
While effective for deterministic models that lack intrinsic sources of randomness, modern machine learning models are inherently stochastic, with randomness arising from initialization, data shuffling, and optimization dynamics. 
Further, bootstrapping is computationally intensive and thus often not economically sensible; \citet{riley2023stability} recommend the training of at least 200 models.
Empirically, \citet{lakshminarayanan2017simple} found that training each model on the entire dataset---using random initialization and data shuffling---achieved better performance than bootstrapping. 
This may be even more pronounced in medical settings with rare and underrepresented conditions.
In summary, while bootstrapping is conceptually simple, ensembling full datasets is a more practical and effective approach for deep learning models with inherent stochasticity, especially in medical settings where data for certain conditions is scarce.
In our experiments (see Section~\ref{subsec:stability}), we found that ensembles of as few as five models with an abstention capability substantially reduce predictive multiplicity while improving accuracy. 

\bmhead{Acknowledgements}
SL received support from the Research and Development Program Information and Communication Technology Bavaria, DIK0444/03. 
KS received support from the German Ministry of Education and Research and the Medical Informatics Initiative as part of the PrivateAIM Project, from the Bavarian Collaborative Research Project PRIPREKI of the Free State of Bavaria Funding Programme "Artificial Intelligence -- Data Science".
This project was funded by the German Ministry of Education and Research under the PrivateAIM Project (reference 01ZZ2316C).
GK conducted this work partially at Google DeepMind.

\bibliography{sn-bibliography}

\end{document}